\documentclass[conference]{IEEEtran}
\usepackage[inner=0.75in,outer=0.75in,top=0.75in,bottom=0.75in]{geometry}

\IEEEoverridecommandlockouts
\usepackage{cite}
\usepackage{amsmath,amssymb,amsfonts}
\usepackage{algorithmic}
\usepackage{graphicx}
\usepackage{textcomp}
\usepackage{xcolor}
\usepackage{float}
\usepackage{soul}

\usepackage{fancyhdr}

\def\BibTeX{{\rm B\kern-.05em{\sc i\kern-.025em b}\kern-.08em
    T\kern-.1667em\lower.7ex\hbox{E}\kern-.125emX}}

\makeatletter 
\newcommand{\linebreakand}{%
  \end{@IEEEauthorhalign}
  \hfill\mbox{}\par
  \mbox{}\hfill\begin{@IEEEauthorhalign}
}
\makeatother 

\begin{document}

\title{Exploring Generative AI for Sim2Real in Driving Data Synthesis\\
\thanks{$^*$ These two authors contributed equally to this work.}
\thanks{Corresponding author: Yiting Wang (yiting.wang.1@warwick.ac.uk). This research is partially sponsored by the Centre for Doctoral
Training to Advance the Deployment of Future Mobility Technologies (CDT FMT) at the University of Warwick.}
}

\author{\IEEEauthorblockN{Haonan Zhao$^*$}
\IEEEauthorblockA{\textit{WMG, University of Warwick} \\
\\
}
\and
\IEEEauthorblockN{Yiting Wang$^*$}
\IEEEauthorblockA{\textit{WMG, University of Warwick} \\
\\
}
\and
\linebreakand
\IEEEauthorblockN{Thomas Bashford-Rogers}
\IEEEauthorblockA{\textit{WMG, University of Warwick} \\
\\
}
\and
\IEEEauthorblockN{Valentina Donzella}
\IEEEauthorblockA{\textit{WMG, University of Warwick} \\
\\
}
\and
\IEEEauthorblockN{Kurt Debattista}
\IEEEauthorblockA{\textit{WMG, University of Warwick} \\
\\
}
}

\maketitle

\thispagestyle{fancy}            
\fancyhead{}                     

\chead{\textit{This work has been submitted to the IEEE for possible publication.} 
}               
\cfoot{\quad}                    
 
\renewcommand{\headrulewidth}{0pt}      
 
\pagestyle{empty}                

\begin{abstract}
Datasets are essential for training and testing vehicle perception algorithms. However, the collection and annotation of real-world images is time-consuming and expensive. 
Driving simulators offer a solution by automatically generating various driving scenarios with corresponding annotations, but the simulation-to-reality (Sim2Real) domain gap remains a challenge.
While most of the Generative Artificial Intelligence (AI) follows the de facto Generative Adversarial Nets (GANs)-based methods, the recent emerging diffusion probabilistic models have not been fully explored in mitigating Sim2Real challenges for driving data synthesis. To explore the performance, this paper applied three different generative AI methods to leverage semantic label maps from a driving simulator as a bridge for the creation of realistic datasets. A comparative analysis of these methods is presented from the perspective of image quality and perception. New synthetic datasets, which include driving images and auto-generated high-quality annotations, are produced with low costs and high scene variability. 
The experimental results show that although GAN-based methods are adept at generating high-quality images when provided with manually annotated labels, ControlNet produces synthetic datasets with fewer artefacts and more structural fidelity when using simulator-generated labels.
This suggests that the diffusion-based approach may provide improved stability and an alternative method for addressing Sim2Real challenges.
These insights contribute to the intelligent vehicle community's understanding of the potential for diffusion models to mitigate the Sim2Real gap.
\end{abstract}


\section{Introduction}
Validation and verification for intelligent vehicles (IVs) heavily rely on driving datasets. To achieve a higher level of automation under various perception tasks, a large number of labelled driving datasets have been collected and released, significantly benefiting the development and deployment of assisted and automated driving (AAD) systems \cite{geiger2012we,yu2020bdd100k,geiger2013vision,cordts2016cityscapes}. However, collecting and annotating real-world data is labour-expensive, not to mention the potential risks of imprecise labelling caused by manual annotation \cite{li2018paralleleye}. Furthermore, it is not feasible for current real-world datasets to encompass all the various driving scenarios due to safety concerns, resulting in a lack of generalisation to unforeseen situations in the development of the perception algorithms. It remains a challenge to get diverse high-quality driving datasets with accurate annotations at a low cost \cite{wang2023effect}.

\begin{figure*}[t!]
\centering
\includegraphics[width=0.92\textwidth]{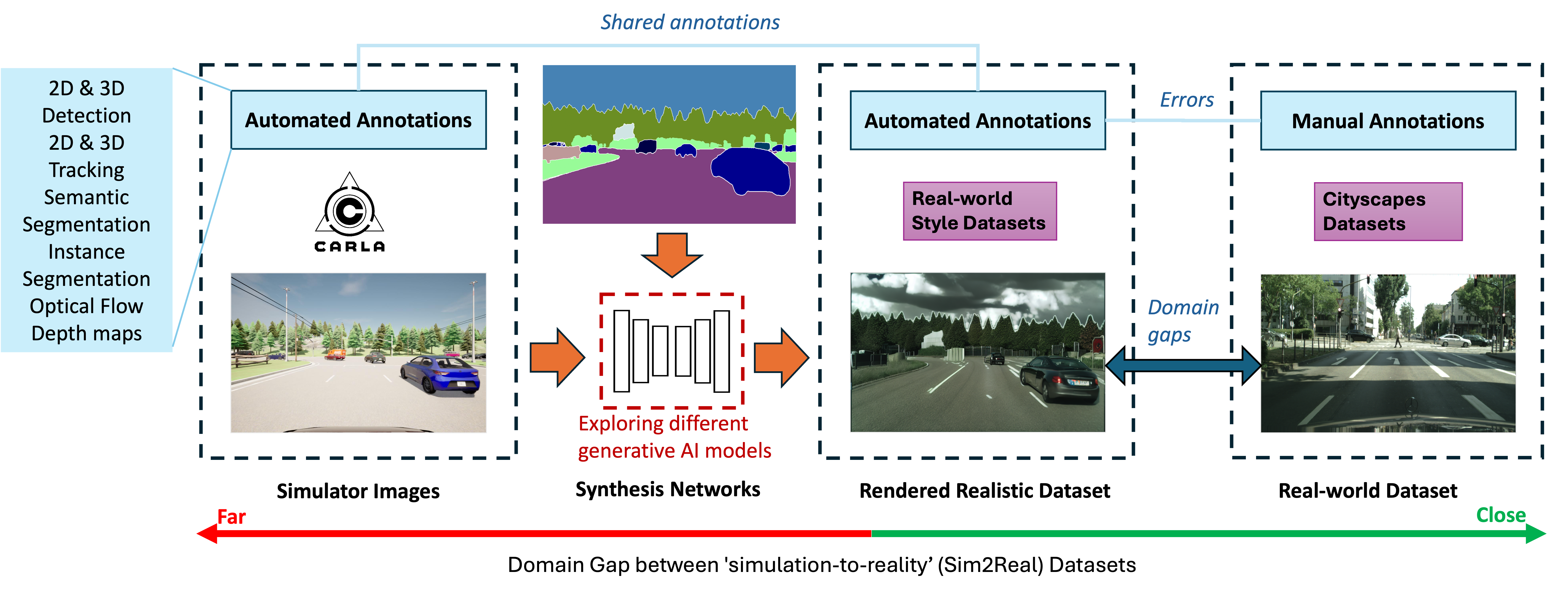}
\caption{The general driving data synthesis pipeline. On the left are the images from the simulator, the right are the generated driving datasets. Different generative AI models are explored as synthetic networks. The domain gap can be reduced after the pipeline. }
\label{pipeline}
\end{figure*}

Given the problems above, an alternative approach is to use the driving simulators to automatically produce diverse driving scenes with corresponding annotations efficiently \cite{kadian2020sim2real,sun2022shift}. For example, the commonly used game engine UNITY is implemented in the ParallelEye pipeline for generating high-quality driving images from simulated images \cite{li2023paralleleye}. 
The driving simulator CARLA was used to synthesise the large-scale driving dataset SHIFT at both discrete and continuous levels to address the domain shift problems \cite{sun2022shift}. This simulator has powerful scenario generation and event editing abilities under different weather conditions for multi-sensor data \cite{sun2022shift}. While they show good flexibility, there exist domain gaps between the simulated data and real-world scenes in terms of realism, see Fig. \ref{pipeline}. One of the reasons for the occurrence of this reality gap is that the synthetic data is not as photo-realistic as real-world data, in terms of the lighting, texture and dynamics\cite{hu2023simulation}. These domain gaps may lead to poor performance, or even invalid models when trained on the simulated dataset if deployed directly in the real world, resulting in potential unsafe behaviour in IVs \cite{wang2023semantic,wang2023effect,wang2024benchmarking}. Therefore, the realism of the driving datasets is an important factor to consider when developing more robust perception algorithms or building the digital twin for IVs \cite{hu2023simulation,zhou2022vetaverse}. 

A number of GAN-based methods have been proposed to reduce the simulation-to-reality (Sim2Real) gap through image-to-image translation \cite{lin2020gan,jung2022exploring,baek2021rethinking,park2020contrastive,kim2022style,torbunov2023uvcgan,schonfeld2023discovering}. 
These Sim2Real methods bridge the reality gap from the synthetic source to the real world by leveraging the semantic label maps. However, these current GAN-based methods show limitations in generating complex scenes and produce results with artefacts \cite{wang2022semantic}. Diffusion models have been proposed as a new class of generative networks that refines performance by generating images via interactive denoising steps and maximum likelihood learning, which have frequently improved performance over GANs in various image generation benchmarks \cite{dhariwal2021diffusion,wang2022semantic,zhang2023adding,nichol2021glide,kim2022diffusionclip,li2024aldm}. Due to the lack of the research utilizing this new types of the generative AI models to explore the potential of the text-guided diffusion-based models in driving data synthesis, this paper explores such methods. 
As illustrated in Fig. \ref{pipeline}, there are three main steps in this exploratory work: simulator image generation, semantic label maps-to-image translation via different generative AI models, and evaluation. The CARLA simulator is utilised for the data generation and three generative AI models are trained using the semantic label maps and the Cityscapes validation sets in a supervised manner. These three generative AI models consist of two GAN-based approaches and a recent diffusion method. 
The performance using these different models is thoroughly evaluated using both image quality metrics and perception scores. Therefore, ours is the first work that explores the potential of various generative AI methods for driving data synthesis based on both real-world and simulated information. 

The main contributions of this work can be summarised as:
\begin{itemize}
\item{The exploration of both GAN and the diffusion-based approaches in a general pipeline to produce large-scale photo-realistic driving datasets based on semantic label maps from simulator and real-world datasets.}
\item{New synthetic datasets based on existing SHIFT datasets, which are suitable for the validation of multiple perception tasks.}
\item{The evaluation of the three methods using both image quality metrics and perception scores.}
\end{itemize}

\section{Related Work}
This section reviews image-to-image (I2I) methods using GAN-based models, diffusion-based models and Sim2Real methods for AAD application.

\textbf{GAN models.} Generative adversarial networks are one of the most widely used approaches for image-to-image translation with its strong ability via generators and discriminators \cite{zhu2017unpaired, park2020contrastive, baek2021rethinking,yang2022unsupervised,jung2022exploring,kim2022style}. One category is the unpaired I2I translation method, which is widely used with the advantage of not requiring paired datasets during training. For example, the widely used CycleGAN introduced a cycle-consistency loss to make sure the generated images are similar to the input images after two rounds of translations - one from A to B and the other from B to A \cite{zhu2017unpaired}. 
As for paired I2I translation methods, many image synthesis approaches have been proposed to generate realistic images based on their paired layout label images, such as semantic layouts or edge maps. Pix2Pix \cite{isola2017image} used a U-Net generator and a patch-based discriminator to build the mapping relationship between layout labels and real photos. 
OASIS \cite{sushko2022oasis} re-designed the discriminator as a segmentation network to maintain semantic information. Ctrl-SIS \cite{schonfeld2023discovering} used a set of class-specific latent directions to guide the network to generate specific local information such as structure or texture. However, most existing GAN models suffer from a lack of generalisation.

\textbf{Diffusion models.} Diffusion models generate images by reversing the process of adding noise to the images via stepwise learning \cite{sohl2015deep}. For example, the latent diffusion models are implemented in various tasks such as image-to-image synthesis \cite{rombach2022high} or text-to-image synthesis \cite{nichol2021glide,avrahami2022blended,gafni2022make,kim2022diffusionclip,avrahami2023blended,wang2023drivedreamer}.
For instance, \cite{rombach2022high} implemented the diffusion process in the latent space of pre-trained autoencoders to reduce the computational cost. The pre-trained language models, such as CLIP \cite{radford2021learning}, are frequently used for generating more diverse results. 
However, the text-guided diffusion-based generation methods provide little control for the generated content. Therefore, various diffusion models have been proposed that offer various controllability to the generated content. For instance, Make-A-Scene \cite{gafni2022make} introduced semantic layouts as additional inputs to the text-to-image diffusion model, which improved the quality of image structure. ControlNet provides control for image structure, style and texture with guidance from control images such as semantic layouts, depth maps and canny edges \cite{zhang2023adding}. 

\textbf{Sim-to-Real in Automated Driving.}
Although a large number of labelled driving datasets have already been collected and released, conducting driving tests via simulators can significantly benefit the data diversity, and provide high flexibility with a reduced cost \cite{hu2023simulation}. Motivated by this, several synthetic driving datasets based on graphics or game engines have been proposed. For instance, GTA-V \cite{richter2016playing} is a synthetic dataset collected from a video game, while SHIFT \cite{sun2022shift} was produced through a professional driving simulator, CARLA \cite{Dosovitskiy17}. \textit{However, as lighting, texture and dynamics in the simulator can be different from the real world, there exist domain gaps between the simulator and real-world data.} To bridge this gap, many Sim2Real approaches have been proposed. Most approaches focus on transferring the knowledge for perception tasks learned by networks from simulator data to reality. For instance, knowledge distillation is used in \cite{hinton2015distilling}, so that the knowledge learned from the simulator data in teacher models can be transferred to the student models for real-world applications. Meta-learning techniques, as described by Hospedales et al. \cite{hospedales2021meta}, equip networks with the ability to `learn to learn' enhancing their adaptability to real-world conditions using experience from simulated tasks. Similarly, Li et. al. minimize the Sim2Real domain gap by stylistically aligning synthetic data with real-world imagery to  \cite{li2018paralleleye}. 
While existing Sim2Real techniques have laid the groundwork for narrowing the domain gaps, challenges remain, particularly in the generation of high-resolution, artefact-free images that are critical for advanced automated driving (AAD) applications. 


\section{Methodology}
As illustrated in Fig. \ref{pipeline}, this paper explores different generative AI approaches (i.e. image-to-image and text-to-image models), to enhance the Sim2Real process. The aim is to utilise the semantic label maps for generating a large scale of annotated photo-realistic driving datasets at a low cost. Please note that other conditional image formats such as depth maps and edge maps can also be used as inputs. However, based on our experimental results, the semantic labels show the best performance. In addition, most of the previous I2I methods use semantic maps, therefore in this paper, we only discuss using semantic maps as the input.


This involves three steps: label-to-image translation model training, simulated to realistic dataset translation and evaluation. After training networks with clean real-world images, two sets of images are generated using the semantic label maps from the real world and the semantic label maps from the simulator. 
We define the semantic label maps in real-world datasets as $S_{real}$ and these in simulated datasets as $S_{sim}$. The photos in real-world datasets are assumed as $V_{real}$. If the generative AI model is expressed as $\phi(\cdot)$, real-world style images generated from $S_{real}$ can be denoted as $\phi (S_{real})$ while these generated from $S_{sim}$ can be denoted as $\phi (S_{sim})$. The overall aim of this work is to explore the performance of different frameworks of $\phi(\cdot)$ on reducing the domain gap between the two synthetic sets ($\phi (S_{real})$, $\phi (S_{sim})$) and $V_{real}$.

\subsection{Selected generative AI models.}
Three generative AI models are chosen to explore their different performances. These are the GAN-based image-to-image translation methods Pix2pixHD \cite{wang2018high} $\phi_{p}(\cdot)$ and OASIS \cite{sushko2022oasis} $\phi_{o}(\cdot)$ and the diffusion-based text-to-image translation method ControlNet $\phi_{c}(\cdot)$. They are chosen due to their representative state-of-the-art performance as discussed below.

Pix2pixHD \cite{wang2018high} is a generative adversarial framework for synthesizing high-resolution images from semantic and instance label maps. It includes a coarse-to-fine generator, a local enhancer and multi-scale discriminators, which maintain both the global and the local information of the images. In addition, the instance maps are combined with the semantic maps to maintain more detailed object boundary information. OASIS \cite{sushko2022oasis} concatenates noise with semantic label maps and a semantic segmentation network into the discriminator to enable a better semantic learning ability of the networks. These two GAN-based approaches are chosen in this work since they achieve the best scores in terms of image quality in layout-to-image tasks according to previous research \cite{li2024aldm}.
 
The main difference between this work with the previous pipeline in \cite{li2023paralleleye} is the consideration of the text-guided image-to-image translation methods, where the popular ControlNet is used \cite{zhang2023adding}. It has the advantage of leveraging information from the large pre-trained stable diffusion model while controlling part of its trainable copy to maintain the image layout information. In addition, this network has a higher level of flexibility with the ability to take text as inputs to guide the image translation process. Since the previous research has often demonstrated the superiority of diffusion-based approaches against GAN-based approaches in terms of realism and image quality \cite{li2023gligen} in image generation tasks, we investigate whether this approach performs well on the driving data synthesis task shown in Fig. \ref{pipeline}.

\subsection{Evaluation}
\textit{\textbf{1)} Image quality:} two sets of image quality metrics are applied, which include the reference-based metrics (i.e. PSNR \cite{PSNR}, SSIM \cite{wang2003multiscale}, LPIPS \cite{zhang2018unreasonable}, CW-SSIM \cite{sampat2009complex}, FSIM \cite{zhang2011fsim}) and non-reference-based image quality metrics (i.e. BRISQUE \cite{mittal2012no}, NIQE \cite{mittal2012making}, FID \cite{heusel2017gans}). The metrics are chosen as they cover both spatial and frequency-based image evaluation. 

\textit{\textbf{2)} Perception:} three frequently used pixel-level perception tasks (i.e. instance segmentation, semantic segmentation and panoptic segmentation) are used to evaluate the performance of the images generated from Cityscapes semantic label maps, which are average precision (AP), panoptic quality (PQ), mean Intersection over Union (mIoU) metrics. 
Higher scores of these metrics indicate better alignment of the generated images with the ground truth datasets. To predict segmentation results using the generated images, this paper applies Oneformer \cite{jain2023oneformer}, a network leveraging multi-task universal structure and a task-conditioned training strategy with state-of-the-art segmentation performance.

\section{Experiment Results}

\begin{figure}[t!]
\centering
\includegraphics[width= 0.5\textwidth]{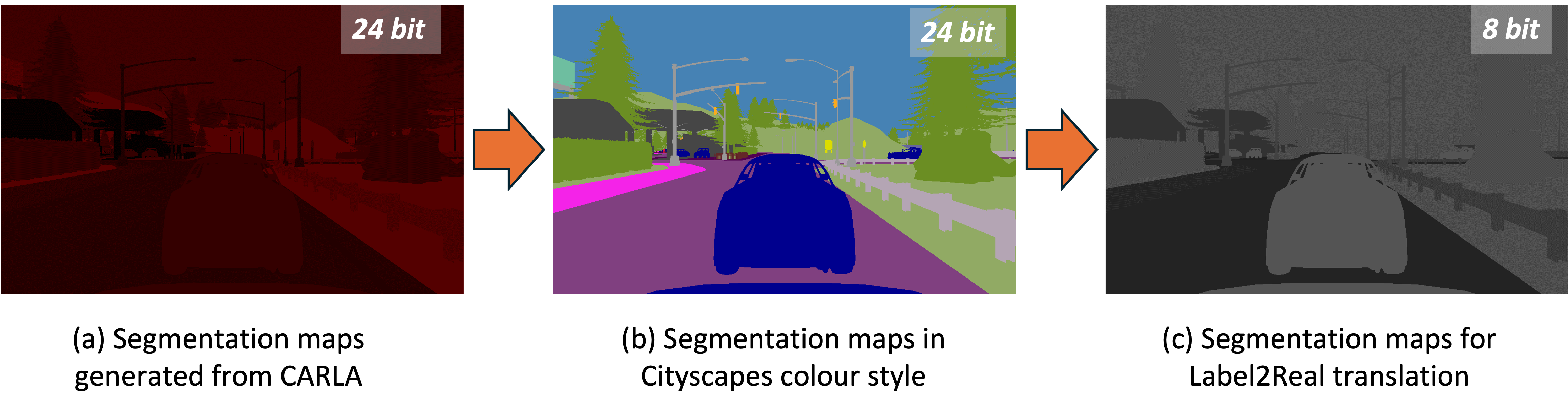}
\caption{The processing pipeline of the different label maps from the CARLA simulator to the Cityscapes style for training. Please note that the picture in (a) is coordinated for better visualisation purposes. }
\label{labels}
\vspace{-5pt}
\end{figure}

\subsection{Implementation Details}
\textbf{Training Details.} We train all our models on an NVIDIA Quadro RTX 8000 GPU with 48GB of GPU memory. In the training process, OASIS was trained with 75 epochs (102 GPU hours). ControlNet was trained for 152 epochs, 210 GPU hours, with a single simple prompt `A realistic driving scene'. Since the released pre-trained Pix2pixHD model has already been trained on high-resolution Cityscapes with 200 epochs, it is simply applied to our experiments. Please note that the number of epochs was chosen based on convergence. 

\textbf{Datasets.} This research resizes all the images into a resolution of $1024\times512$ during all the experiments since a higher resolution is necessary for high-quality driving datasets \cite{cordts2016cityscapes,geiger2013vision,yu2020bdd100k}.
\textit{\textbf{1)} Training Set.} Cityscapes is chosen in our experiments since it is a widely used urban driving dataset captured with high quality; it also contains labels that are required for our experiments. Specifically, 2975 images are used for training in this paper. \textit{ \textbf{2)} Validation Set.} For evaluation, all 500 images from the Cityscapes were used ($V_i$), and three new validation sets were generated from the semantic label maps from Cityscapes via OASIS ($V_{oc}$), Pix2pixHD ($V_{pc}$) and ControlNet ($V_{cc}$), respectively (each has 500 images). The validation set from the synthetic dataset SHIFT is also used for the comparison since it is the dataset generated directly from the CARLA simulator with all the required labels \cite{sun2022shift}. We generate three validation sets via OASIS ($V_{os}$), Pix2pixHD ($V_{ps}$) and ControlNet ($V_{cs}$), respectively, from the SHIFT validation set. Since the validation set of SHIFT contains 25500 images (500 scenarios, each with 51 images), one image is randomly selected from each scenario to make a fairer comparison with more diversity. As a result, a validation set, which contains 500 images of highway, suburban, and village conditions from the CARLA simulator, is used in experiments. Due to the different format, the 24-bit-depth semantic label maps from CARLA are converted to 8-bit-depth to match Cityscapes labels, see Fig. \ref{labels}.

\subsection{Results and Analysis}

\begin{figure*}[t!]
\centering
\includegraphics[width= 1\textwidth]{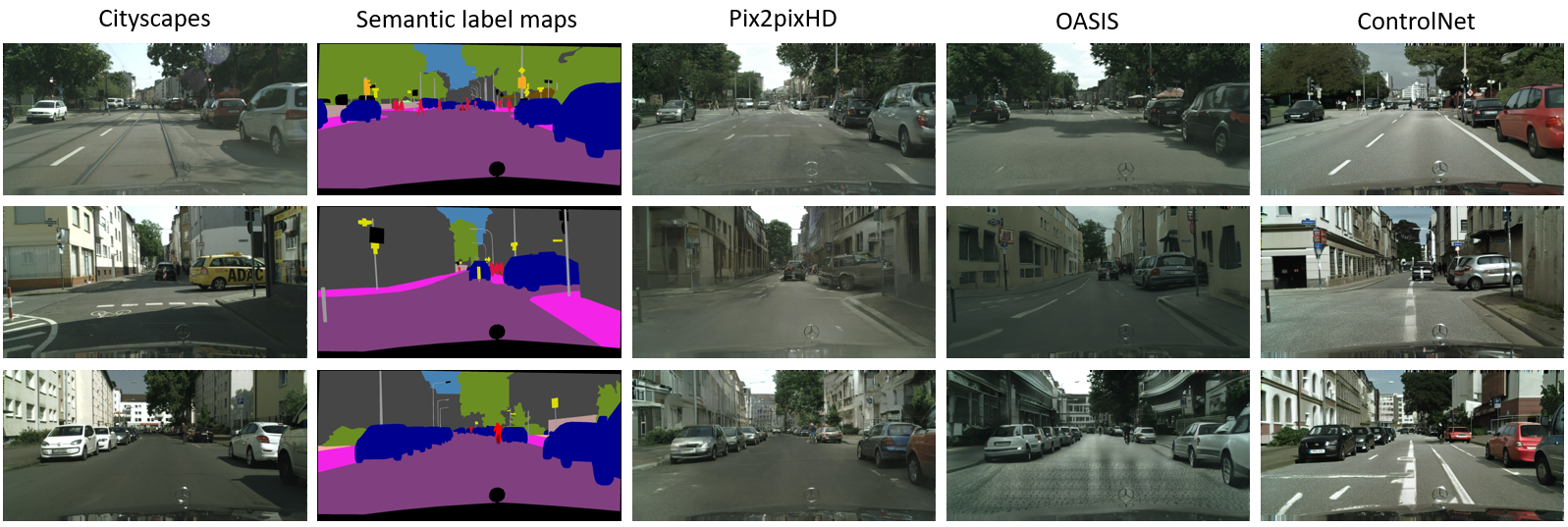}
\caption{Qualitative comparison of results from different generative approaches with the Cityscapes validation set}
\label{visual_results}
\vspace{-5pt}
\end{figure*}

\begin{figure*}[t!]
\centering
\includegraphics[width= 1\textwidth]{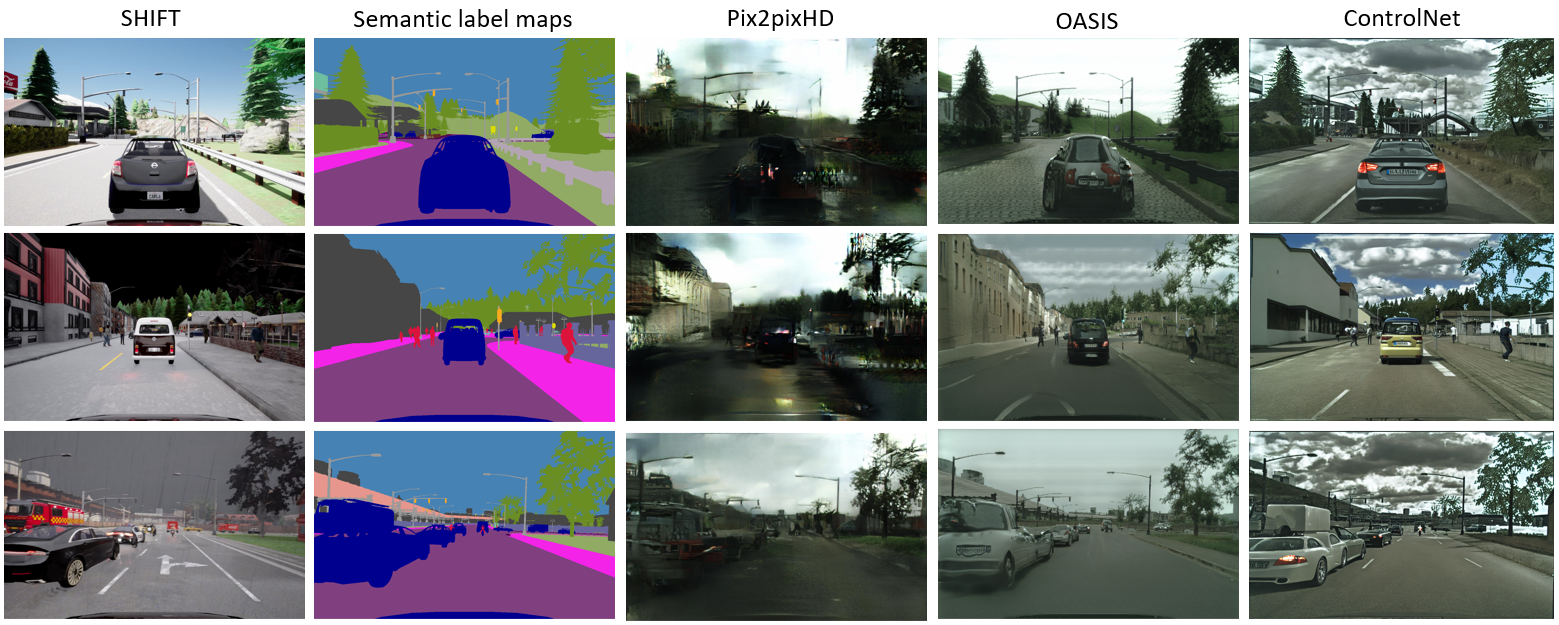}
\caption{Qualitative comparison of results from different generative approaches with the SHIFT validation set}
\label{visual_results_2}
\vspace{-5pt}
\end{figure*}

\begin{table}[t!]
\centering
\caption{ Evaluation results for image quality and fidelity on three cityscapes-based synthetic validation sets. \textit{↑ and ↓ indicate that the higher or lower the scores, the better or worse the performance.}}
\begin{tabular}{cccc}
\hline
        \textbf{{\begin{tabular}[c]{@{}c@{}}Data\\ (method)\end{tabular}}}
        & {{\begin{tabular}[c]{@{}c@{}}$V_{pc}$\\ (Pix2pixHD)\end{tabular}}} 
        & {{\begin{tabular}[c]{@{}c@{}}$V_{oc}$\\ (OASIS)\end{tabular}}}  
        & {{\begin{tabular}[c]{@{}c@{}}$V_{cc}$\\ (ControlNet)\end{tabular}}} 
        
        \\ \hline
\textbf{PSNR↑}    & \color{red}16.142    & 14.816 & \color{blue}12.461     \\
\textbf{SSIM↑}    & \color{red}0.484     & 0.451  & \color{blue}0.334      \\
\textbf{LPIPS↓}   & \color{red}0.478     & 0.510  & \color{blue}0.536      \\
\textbf{CW-SSIM↑} & \color{red}0.651     & 0.631  & \color{blue}0.555      \\
\textbf{FSIM↑}    & \color{red}0.676     & 0.650  & \color{blue}0.598      \\
\textbf{NIQE↓}    & 12.908    & \color{red}12.416 & \color{blue}12.929     \\
\textbf{BRISQUE↓} & \color{blue}31.471    & 26.126 & \color{red}22.238     \\
\textbf{FID↓}     & 57.477    & \color{red}52.394 & \color{blue}65.264     \\ \hline
\end{tabular}
\label{iq}
\end{table}

\begin{table}[t!]
\centering
\caption{evaluation results for perception on three cityscapes-based synthetic validation sets. The pre-trained oneformer model is chosen as the segmentation model.}
\begin{tabular}{lccc}
\hline

         \textbf{Data (method)} & \textbf{PQ↑}           & \textbf{mIoU↑}              & \textbf{AP↑}              \\ \hline 
$V_{i}$ (Baseline) & 68.5                        & 83.0                        & 46.5   \\ 
$V_{pc}$(Pix2pixHD)    & 54.7                    & 73.9                    & \color{red}25.9             \\
$V_{oc}$(OASIS)        & \color{red}60.0       & \color{red}77.0           & 25.4                        \\
$V_{cc}$(ControlNet)   & \color{blue}42.5      & \color{blue}61.9          & \color{blue}11.5                        \\ \hline

\end{tabular}
\label{pq}
\end{table}

\begin{table}[t!]
\centering
\caption{evaluation results for image quality, fidelity and perception performance on three shift-based synthetic validation sets \textit{(Average values)}.}
\begin{tabular}{cccc|c}
\hline
           & \textbf{FID↓}     & \textbf{BRISQUE↓} & \textbf{NIQE↓} & \textbf{mIoU↑}  \\ \hline
Pix2pixHD($V_{ps}$)  & \color{blue}162.690 & \color{blue}43.447  & \color{blue}13.183    & \color{blue}18.8   \\
OASIS($V_{os}$)      & \color{red}112.405 & 31.899  & 10.780     &  33.5   \\
ControlNet($V_{cs}$) & 131.423 & \color{red}14.625  & \color{red}9.175  &  \color{red}34.1   \\ \hline
\end{tabular}
\label{shift1}
\end{table}

\textbf{Quantitative Results.} \textit{1) Synthetic results based on Cityscapes semantic label maps.}
The image quality evaluation results between the original Cityscapes validation set $V_{i}$ and the three synthetic validation sets $V_{pc}$, $V_{oc}$ and $V_{cc}$ are shown in Table \ref{iq}. It can be seen that the GAN-based approaches achieve better label-to-image translation results compared with the diffusion-based method in general. Pix2pixHD achieves the best score in all the reference-based metrics, from both the spatial and frequency domains. OASIS has the best naturalness in terms of the NIQE and the FID, while ControlNet achieves the best BRISQUE score.
It can be indicated from FID scores that the images in $V_{oc}$ are the closest to the target real images while the images in $V_{cc}$ are far from them, which shows the results of ControlNet have less realism than real-world images. LPIPS scores illustrate that the similarity between ControlNet results and real Cityscapes images is the least according to human perception. NIQE scores show that $V_{cc}$ still has the lowest image quality when evaluated by the no-reference metric.
However, $V_{cc}$ shows the best performance when evaluated by another no-reference metric, BRISQUE.
Table \ref{pq} shows the perception performance of the four different validation datasets: $V_{i}$, $V_{pc}$, $V_{oc}$, and $V_{cc}$. The state-of-the-art segmentation model called Oneformer \cite{jain2023oneformer} is used for this evaluation. The table shows a similar trend, where the data generated by OASIS achieves the best scores in segmentation tasks and ControlNet the lowest.
The architecture of ControlNet mainly consists of the large diffusion model concatenated with its trainable copy. In addition, the weights of the stable diffusion model, the base of ControlNet, are kept locked when the ControlNet is being trained by Cityscapes data to maintain the learned knowledge within the diffusion model. This means that the content generated by ControlNet is not only determined by the images in the training set but also by the previous data used to train the diffusion model. This explains why the $V_{cc}$ set contains content that is unseen in the training set, leading to the least similarity to the Cityscapes dataset.

\textit{2) Synthetic results based on SHIFT semantic label maps.} The results of the image quality of three synthetic datasets based on SHIFT semantic label maps $V_{ps}$, $V_{os}$, and $V_{cs}$ are shown in Tab. \ref{shift1}. $V_{ps}$, $V_{os}$ and $V_{cs}$ are three synthetic validation sets generated by Pix2pixHD, OASIS and ControlNet, respectively, while taking semantic label maps from the SHIFT synthetic dataset as inputs. Since there are no real-world images in the SHIFT dataset for reference, only non-reference-based image quality metrics are calculated by evaluating the distance between real images in Cityscapes and three synthetic validation sets. Also, because annotations for panoptic segmentation are not provided in SHIFT dataset, only mIoU scores are calculated here for the evaluation of perception performance. Although the format of input segmentation label maps (e.g., the colour for each class, the bit depth of maps, see Fig. \ref{labels}) for each network is converted exactly the same as that in the Cityscapes dataset, the results are quite different. The results indicate that the data generated by ControlNet has better naturalness in terms of the BRISQUE and NIQE, while the data generated by OASIS has more similar features as Cityscapes in terms of FID. The mIoU scores show the data generated by ControlNet also achieves the best semantic segmentation performance. The potential reason could be that the label maps in Cityscapes are annotated manually and may have some errors, while the label maps from SHIFT are automatically generated by the simulator with pixel-level accuracy.

\textbf{Qualitative Results.}
\textit{ 1) Visual results based on Cityscapes semantic label maps.} Fig. \ref{visual_results} shows the visual comparison of $V_{pc}$, $V_{oc}$ and $V_{cc}$ using the Cityscapes labels as input. The preservation of edge information across all results is evident, supported by the semantic label maps. Although ControlNet's output stylistically diverges from the Cityscapes dataset images, the clarity and detail within the objects are markedly improved, and the occurrence of artefacts is minimized. In contrast, the two GAN-based methods, Pix2pixHD and OASIS, demonstrate issues with distortion and blurriness. Notably, OASIS-produced images are sometimes marred by black checkerboard patterns on the road surfaces. 

\textit{ 2) Visual results based on SHIFT semantic label maps.} Fig. \ref{visual_results_2} shows a qualitative comparison of $V_{ps}$, $V_{os}$ and $V_{cs}$ generated from the CARLA simulator labels. It can be seen that the Pix2PixHD has the worst satisfying performance, with the majority of the areas blurry and distorted with black artefacts, and the overall structure of the entire image lacks meaningful coherence. Images generated by OASIS keep a similar style to real images, but there exists a considerable amount of distortion. As indicated in the third row in Fig. \ref{visual_results_2}, the GAN-based methods cannot understand the occlusion relationship of the vehicles (e.g. it appears to misunderstand the occluded big truck as part of the cars), while the results generated by the ControlNet accurately display this occlusion.  

\section{Discussion}
Overall, it can be seen from all the quantitative results that the data generated by ControlNet fails to perform well on image quality and perception tasks when using Cityscapes labels as inputs. The reason could be that the two GAN-based approaches train networks through well-designed adversarial loss that can supervise the outputs to approximate the real images by incorporating a feature matching loss (Pix2pixHD) or applying a content-aware regularization (OASIS) \cite{wang2018high, sushko2022oasis}. Pix2pixHD not only considers the semantic but also the instance label maps, which maintain better object boundary information. Also, the discriminator in the OASIS network is a segmentation network designed specifically to enhance segmentation performance. These designs facilitate the two GAN-based approaches to generate images similar to the Cityscapes style and perform well on segmentation tasks. In contrast, the design of the ControlNet is to add prompts and conditions to control the synthesis results through its concatenation with the large pre-trained stable diffusion model, whose knowledge can also influence the results. Although the two GAN-based approaches show good quantitative performance when taking label maps from Cityscapes as inputs, they lack robustness. Subtle changes in the labels can have significant impacts on their outcomes, resulting in better BRISQUE, NIQE and mIoU scores in data generated from ControlNet when using SHIFT labels as inputs. The main difference between the input semantic label maps from Cityscapes and SHIFT is whether the edges of their labels are pixel-level accurate or not, and it seems that Pix2pixHD cannot generalise this difference. 

The qualitative results indicate a different trend. Although the style of images from ControlNet is less similar to real images in Cityscapes (e.g. brightness, cloudy sky), ControlNet still synthesises the best textures, colours and shapes. In contrast, the results of the two GAN-based methods exhibit blur, distortion, and artefacts when using the semantic labels from the simulator. The potential reason may be that the loss function does not adequately penalize high-frequency errors. Checkerboard artefacts, specifically, are typically a consequence of deconvolutional layers used in GANs that can produce an uneven overlap of the kernels when upsampling, leading to a grid-like pattern on the generated images. ControlNet's resilience to these visual anomalies might be due to its diffusion-based generative process, which iteratively refines the image quality, allowing it to better handle the nuances of complex textures and patterns without introducing high-frequency distortions. In addition, all models fail to keep the same colour information as the original images generated by CARLA; this is to be expected as only the semantic maps are used as input and they do not contain much information which can be used to generate consistent colours.

\section{Conclusion}
In this paper, different generative AI approaches are explored in driving data synthesis to reduce the Sim2Real domain gap. Synthetic validation datasets are generated through three generative approaches and two sources of semantic label maps. Both the image quality metrics and segmentation scores are used for evaluation. Findings reveal that ControlNet shows better generalisation and stability (e.g. structural information) in driving data synthesis when using labels from different sources (i.e. simulator). It may be due to the extra guidance from the prompts. In the future, the modification of the ControlNet and better evaluation methods can be potentially explored to improve the diversity and realism of generated images.

\bibliographystyle{IEEEtran}
\bibliography{egbib}

\begin{thebibliography}{10}
\providecommand{\url}[1]{#1}
\csname url@samestyle\endcsname
\providecommand{\newblock}{\relax}
\providecommand{\bibinfo}[2]{#2}
\providecommand{\BIBentrySTDinterwordspacing}{\spaceskip=0pt\relax}
\providecommand{\BIBentryALTinterwordstretchfactor}{4}
\providecommand{\BIBentryALTinterwordspacing}{\spaceskip=\fontdimen2\font plus
\BIBentryALTinterwordstretchfactor\fontdimen3\font minus \fontdimen4\font\relax}
\providecommand{\BIBforeignlanguage}[2]{{%
\expandafter\ifx\csname l@#1\endcsname\relax
\typeout{** WARNING: IEEEtran.bst: No hyphenation pattern has been}%
\typeout{** loaded for the language `#1'. Using the pattern for}%
\typeout{** the default language instead.}%
\else
\language=\csname l@#1\endcsname
\fi
#2}}
\providecommand{\BIBdecl}{\relax}
\BIBdecl

\bibitem{geiger2012we}
A.~Geiger, P.~Lenz, and R.~Urtasun, ``Are we ready for autonomous driving? the kitti vision benchmark suite,'' in \emph{2012 IEEE conference on computer vision and pattern recognition}.\hskip 1em plus 0.5em minus 0.4em\relax IEEE, 2012, pp. 3354--3361.

\bibitem{yu2020bdd100k}
F.~Yu, H.~Chen, X.~Wang, W.~Xian, Y.~Chen, F.~Liu, V.~Madhavan, and T.~Darrell, ``Bdd100k: A diverse driving dataset for heterogeneous multitask learning,'' in \emph{Proceedings of the IEEE/CVF conference on computer vision and pattern recognition}, 2020, pp. 2636--2645.

\bibitem{geiger2013vision}
A.~Geiger, P.~Lenz, C.~Stiller, and R.~Urtasun, ``Vision meets robotics: The kitti dataset,'' \emph{The International Journal of Robotics Research}, vol.~32, no.~11, pp. 1231--1237, 2013.

\bibitem{cordts2016cityscapes}
M.~Cordts, M.~Omran, S.~Ramos, T.~Rehfeld, M.~Enzweiler, R.~Benenson, U.~Franke, S.~Roth, and B.~Schiele, ``The cityscapes dataset for semantic urban scene understanding,'' in \emph{Proceedings of the IEEE conference on computer vision and pattern recognition}, 2016, pp. 3213--3223.

\bibitem{li2018paralleleye}
X.~Li, K.~Wang, Y.~Tian, L.~Yan, F.~Deng, and F.-Y. Wang, ``The paralleleye dataset: A large collection of virtual images for traffic vision research,'' \emph{IEEE Transactions on Intelligent Transportation Systems}, vol.~20, no.~6, pp. 2072--2084, 2018.

\bibitem{wang2023effect}
Y.~Wang, H.~Zhao, K.~Debattista, and V.~Donzella, ``The effect of camera data degradation factors on panoptic segmentation for automated driving,'' \emph{26th IEEE International Conference on Intelligent Transportation Systems}, 2023.

\bibitem{kadian2020sim2real}
A.~Kadian, J.~Truong, A.~Gokaslan, A.~Clegg, E.~Wijmans, S.~Lee, M.~Savva, S.~Chernova, and D.~Batra, ``Sim2real predictivity: Does evaluation in simulation predict real-world performance?'' \emph{IEEE Robotics and Automation Letters}, vol.~5, no.~4, pp. 6670--6677, 2020.

\bibitem{sun2022shift}
T.~Sun, M.~Segu, J.~Postels, Y.~Wang, L.~Van~Gool, B.~Schiele, F.~Tombari, and F.~Yu, ``Shift: a synthetic driving dataset for continuous multi-task domain adaptation,'' in \emph{Proceedings of the IEEE/CVF Conference on Computer Vision and Pattern Recognition}, 2022, pp. 21\,371--21\,382.

\bibitem{li2023paralleleye}
X.~Li, K.~Wang, X.~Gu, F.~Deng, and F.-Y. Wang, ``Paralleleye pipeline: An effective method to synthesize images for improving the visual intelligence of intelligent vehicles,'' \emph{IEEE Transactions on Systems, Man, and Cybernetics: Systems}, 2023.

\bibitem{hu2023simulation}
X.~Hu, S.~Li, T.~Huang, B.~Tang, R.~Huai, and L.~Chen, ``How simulation helps autonomous driving: A survey of sim2real, digital twins, and parallel intelligence,'' \emph{IEEE Transactions on Intelligent Vehicles}, 2023.

\bibitem{wang2023semantic}
Y.~Wang, P.~H. Chan, and V.~Donzella, ``Semantic-aware video compression for automotive cameras,'' \emph{IEEE Transactions on Intelligent Vehicles}, 2023.

\bibitem{wang2024benchmarking}
Y.~Wang, H.~Zhao, D.~Gummadi, M.~Dianati, K.~Debattista, and V.~Donzella, ``Benchmarking the robustness of panoptic segmentation for automated driving,'' \emph{arXiv preprint arXiv:2402.15469}, 2024.

\bibitem{zhou2022vetaverse}
P.~Zhou, J.~Zhu, Y.~Wang, Y.~Lu, Z.~Wei, H.~Shi, Y.~Ding, Y.~Gao, Q.~Huang, Y.~Shi \emph{et~al.}, ``Vetaverse: Technologies, applications, and visions toward the intersection of metaverse, vehicles, and transportation systems,'' \emph{arXiv preprint arXiv:2210.15109}, 2022.

\bibitem{lin2020gan}
C.-T. Lin, S.-W. Huang, Y.-Y. Wu, and S.-H. Lai, ``Gan-based day-to-night image style transfer for nighttime vehicle detection,'' \emph{IEEE Transactions on Intelligent Transportation Systems}, vol.~22, no.~2, pp. 951--963, 2020.

\bibitem{jung2022exploring}
C.~Jung, G.~Kwon, and J.~C. Ye, ``Exploring patch-wise semantic relation for contrastive learning in image-to-image translation tasks,'' in \emph{Proceedings of the IEEE/CVF conference on computer vision and pattern recognition}, 2022, pp. 18\,260--18\,269.

\bibitem{baek2021rethinking}
K.~Baek, Y.~Choi, Y.~Uh, J.~Yoo, and H.~Shim, ``Rethinking the truly unsupervised image-to-image translation,'' in \emph{Proceedings of the IEEE/CVF International Conference on Computer Vision}, 2021, pp. 14\,154--14\,163.

\bibitem{park2020contrastive}
T.~Park, A.~A. Efros, R.~Zhang, and J.-Y. Zhu, ``Contrastive learning for unpaired image-to-image translation,'' in \emph{Computer Vision--ECCV 2020: 16th European Conference, Glasgow, UK, August 23--28, 2020, Proceedings, Part IX 16}.\hskip 1em plus 0.5em minus 0.4em\relax Springer, 2020, pp. 319--345.

\bibitem{kim2022style}
K.~Kim, S.~Park, E.~Jeon, T.~Kim, and D.~Kim, ``A style-aware discriminator for controllable image translation,'' in \emph{Proceedings of the IEEE/CVF conference on computer vision and pattern recognition}, 2022, pp. 18\,239--18\,248.

\bibitem{torbunov2023uvcgan}
D.~Torbunov, Y.~Huang, H.~Yu, J.~Huang, S.~Yoo, M.~Lin, B.~Viren, and Y.~Ren, ``Uvcgan: Unet vision transformer cycle-consistent gan for unpaired image-to-image translation,'' in \emph{Proceedings of the IEEE/CVF Winter Conference on Applications of Computer Vision}, 2023, pp. 702--712.

\bibitem{schonfeld2023discovering}
E.~Sch{\"o}nfeld, J.~Borges, V.~Sushko, B.~Schiele, and A.~Khoreva, ``Discovering class-specific gan controls for semantic image synthesis,'' in \emph{Proceedings of the IEEE/CVF Conference on Computer Vision and Pattern Recognition}, 2023, pp. 688--697.

\bibitem{wang2022semantic}
W.~Wang, J.~Bao, W.~Zhou, D.~Chen, D.~Chen, L.~Yuan, and H.~Li, ``Semantic image synthesis via diffusion models,'' \emph{arXiv preprint arXiv:2207.00050}, 2022.

\bibitem{dhariwal2021diffusion}
P.~Dhariwal and A.~Nichol, ``Diffusion models beat gans on image synthesis,'' \emph{Advances in neural information processing systems}, vol.~34, pp. 8780--8794, 2021.

\bibitem{zhang2023adding}
L.~Zhang, A.~Rao, and M.~Agrawala, ``Adding conditional control to text-to-image diffusion models,'' in \emph{Proceedings of the IEEE/CVF International Conference on Computer Vision}, 2023, pp. 3836--3847.

\bibitem{nichol2021glide}
A.~Nichol, P.~Dhariwal, A.~Ramesh, P.~Shyam, P.~Mishkin, B.~McGrew, I.~Sutskever, and M.~Chen, ``Glide: Towards photorealistic image generation and editing with text-guided diffusion models,'' \emph{arXiv preprint arXiv:2112.10741}, 2021.

\bibitem{kim2022diffusionclip}
G.~Kim, T.~Kwon, and J.~C. Ye, ``Diffusionclip: Text-guided diffusion models for robust image manipulation,'' in \emph{Proceedings of the IEEE/CVF Conference on Computer Vision and Pattern Recognition}, 2022, pp. 2426--2435.

\bibitem{li2024aldm}
Y.~Li, M.~Keuper, D.~Zhang, and A.~Khoreva, ``Adversarial supervision makes layout-to-image diffusion models thrive,'' in \emph{ICLR}, 2024.

\bibitem{zhu2017unpaired}
J.-Y. Zhu, T.~Park, P.~Isola, and A.~A. Efros, ``Unpaired image-to-image translation using cycle-consistent adversarial networks,'' in \emph{Proceedings of the IEEE international conference on computer vision}, 2017, pp. 2223--2232.

\bibitem{yang2022unsupervised}
S.~Yang, L.~Jiang, Z.~Liu, and C.~C. Loy, ``Unsupervised image-to-image translation with generative prior,'' in \emph{Proceedings of the IEEE/CVF conference on computer vision and pattern recognition}, 2022, pp. 18\,332--18\,341.

\bibitem{isola2017image}
P.~Isola, J.-Y. Zhu, T.~Zhou, and A.~A. Efros, ``Image-to-image translation with conditional adversarial networks,'' in \emph{Proceedings of the IEEE conference on computer vision and pattern recognition}, 2017, pp. 1125--1134.

\bibitem{sushko2022oasis}
V.~Sushko, E.~Sch{\"o}nfeld, D.~Zhang, J.~Gall, B.~Schiele, and A.~Khoreva, ``Oasis: only adversarial supervision for semantic image synthesis,'' \emph{International Journal of Computer Vision}, vol. 130, no.~12, pp. 2903--2923, 2022.

\bibitem{sohl2015deep}
J.~Sohl-Dickstein, E.~Weiss, N.~Maheswaranathan, and S.~Ganguli, ``Deep unsupervised learning using nonequilibrium thermodynamics,'' in \emph{International conference on machine learning}.\hskip 1em plus 0.5em minus 0.4em\relax PMLR, 2015, pp. 2256--2265.

\bibitem{rombach2022high}
R.~Rombach, A.~Blattmann, D.~Lorenz, P.~Esser, and B.~Ommer, ``High-resolution image synthesis with latent diffusion models,'' in \emph{Proceedings of the IEEE/CVF conference on computer vision and pattern recognition}, 2022, pp. 10\,684--10\,695.

\bibitem{avrahami2022blended}
O.~Avrahami, D.~Lischinski, and O.~Fried, ``Blended diffusion for text-driven editing of natural images,'' in \emph{Proceedings of the IEEE/CVF Conference on Computer Vision and Pattern Recognition}, 2022, pp. 18\,208--18\,218.

\bibitem{gafni2022make}
O.~Gafni, A.~Polyak, O.~Ashual, S.~Sheynin, D.~Parikh, and Y.~Taigman, ``Make-a-scene: Scene-based text-to-image generation with human priors,'' in \emph{European Conference on Computer Vision}.\hskip 1em plus 0.5em minus 0.4em\relax Springer, 2022, pp. 89--106.

\bibitem{avrahami2023blended}
O.~Avrahami, O.~Fried, and D.~Lischinski, ``Blended latent diffusion,'' \emph{ACM Transactions on Graphics (TOG)}, vol.~42, no.~4, pp. 1--11, 2023.

\bibitem{wang2023drivedreamer}
X.~Wang, Z.~Zhu, G.~Huang, X.~Chen, and J.~Lu, ``Drivedreamer: Towards real-world-driven world models for autonomous driving,'' \emph{arXiv preprint arXiv:2309.09777}, 2023.

\bibitem{radford2021learning}
A.~Radford, J.~W. Kim, C.~Hallacy, A.~Ramesh, G.~Goh, S.~Agarwal, G.~Sastry, A.~Askell, P.~Mishkin, J.~Clark \emph{et~al.}, ``Learning transferable visual models from natural language supervision,'' in \emph{International conference on machine learning}.\hskip 1em plus 0.5em minus 0.4em\relax PMLR, 2021, pp. 8748--8763.

\bibitem{richter2016playing}
S.~R. Richter, V.~Vineet, S.~Roth, and V.~Koltun, ``Playing for data: Ground truth from computer games,'' in \emph{Computer Vision--ECCV 2016: 14th European Conference, Amsterdam, The Netherlands, October 11-14, 2016, Proceedings, Part II 14}.\hskip 1em plus 0.5em minus 0.4em\relax Springer, 2016, pp. 102--118.

\bibitem{Dosovitskiy17}
A.~Dosovitskiy, G.~Ros, F.~Codevilla, A.~Lopez, and V.~Koltun, ``{CARLA}: {An} open urban driving simulator,'' in \emph{Proceedings of the 1st Annual Conference on Robot Learning}, 2017, pp. 1--16.

\bibitem{hinton2015distilling}
G.~Hinton, O.~Vinyals, and J.~Dean, ``Distilling the knowledge in a neural network,'' \emph{arXiv preprint arXiv:1503.02531}, 2015.

\bibitem{hospedales2021meta}
T.~Hospedales, A.~Antoniou, P.~Micaelli, and A.~Storkey, ``Meta-learning in neural networks: A survey,'' \emph{IEEE transactions on pattern analysis and machine intelligence}, vol.~44, no.~9, pp. 5149--5169, 2021.

\bibitem{wang2018high}
T.-C. Wang, M.-Y. Liu, J.-Y. Zhu, A.~Tao, J.~Kautz, and B.~Catanzaro, ``High-resolution image synthesis and semantic manipulation with conditional gans,'' in \emph{Proceedings of the IEEE conference on computer vision and pattern recognition}, 2018, pp. 8798--8807.

\bibitem{li2023gligen}
Y.~Li, H.~Liu, Q.~Wu, F.~Mu, J.~Yang, J.~Gao, C.~Li, and Y.~J. Lee, ``Gligen: Open-set grounded text-to-image generation,'' in \emph{Proceedings of the IEEE/CVF Conference on Computer Vision and Pattern Recognition}, 2023, pp. 22\,511--22\,521.

\bibitem{PSNR}
\BIBentryALTinterwordspacing
Jan 2024. [Online]. Available: \url{https://en.wikipedia.org/wiki/Peak\_signal-to-noise\_ratio}
\BIBentrySTDinterwordspacing

\bibitem{wang2003multiscale}
Z.~Wang, E.~P. Simoncelli, and A.~C. Bovik, ``Multiscale structural similarity for image quality assessment,'' in \emph{The Thrity-Seventh Asilomar Conference on Signals, Systems \& Computers, 2003}, vol.~2.\hskip 1em plus 0.5em minus 0.4em\relax Ieee, 2003, pp. 1398--1402.

\bibitem{zhang2018unreasonable}
R.~Zhang, P.~Isola, A.~A. Efros, E.~Shechtman, and O.~Wang, ``The unreasonable effectiveness of deep features as a perceptual metric,'' in \emph{Proceedings of the IEEE conference on computer vision and pattern recognition}, 2018, pp. 586--595.

\bibitem{sampat2009complex}
M.~P. Sampat, Z.~Wang, S.~Gupta, A.~C. Bovik, and M.~K. Markey, ``Complex wavelet structural similarity: A new image similarity index,'' \emph{IEEE transactions on image processing}, vol.~18, no.~11, pp. 2385--2401, 2009.

\bibitem{zhang2011fsim}
L.~Zhang, L.~Zhang, X.~Mou, and D.~Zhang, ``Fsim: A feature similarity index for image quality assessment,'' \emph{IEEE transactions on Image Processing}, vol.~20, no.~8, pp. 2378--2386, 2011.

\bibitem{mittal2012no}
A.~Mittal, A.~K. Moorthy, and A.~C. Bovik, ``No-reference image quality assessment in the spatial domain,'' \emph{IEEE Transactions on image processing}, vol.~21, no.~12, pp. 4695--4708, 2012.

\bibitem{mittal2012making}
A.~Mittal, R.~Soundararajan, and A.~C. Bovik, ``Making a “completely blind” image quality analyzer,'' \emph{IEEE Signal processing letters}, vol.~20, no.~3, pp. 209--212, 2012.

\bibitem{heusel2017gans}
M.~Heusel, H.~Ramsauer, T.~Unterthiner, B.~Nessler, and S.~Hochreiter, ``Gans trained by a two time-scale update rule converge to a local nash equilibrium,'' \emph{Advances in neural information processing systems}, vol.~30, 2017.

\bibitem{jain2023oneformer}
J.~Jain, J.~Li, M.~T. Chiu, A.~Hassani, N.~Orlov, and H.~Shi, ``Oneformer: One transformer to rule universal image segmentation,'' in \emph{Proceedings of the IEEE/CVF Conference on Computer Vision and Pattern Recognition}, 2023, pp. 2989--2998.

\end{thebibliography}
\end{document}